\documentclass[acmtog, screen, nonacm]{acmart}

\usepackage{booktabs} 
\usepackage{bbm}
\usepackage{nicefrac}
\usepackage{algorithm}
\usepackage{algpseudocode}
\usepackage[normalem]{ulem}
\usepackage{empheq} 
\usepackage{algorithm}
\usepackage{algpseudocode}
\usepackage{hyperref}
\usepackage[percent]{overpic}
\setcopyright{none}
\setcitestyle{square}

\title{DressWild: Feed-Forward Pose-Agnostic Garment Sewing Pattern Generation from In-the-Wild Images}

\author{Zeng Tao}
\authornotemark[1]
\affiliation{
\institution{UCLA and Fudan University}
}

\author{Ying Jiang}
\authornotemark[1]
\affiliation{
\institution{UCLA}
}

\author{Yunuo Chen}
\authornotemark[1]
\affiliation{
\institution{UCLA}
}

\author{Tianyi Xie}
\affiliation{
\institution{UCLA}
}

\author{Huamin Wang}
\affiliation{
\institution{Style3D}
}

\author{Yingnian Wu}
\affiliation{
\institution{UCLA}
}

\author{Yin Yang}
\affiliation{
\institution{University of Utah}
}

\author{Abishek Sampath Kumar}
\affiliation{
\institution{Sony}
}

\author{Kenji Tashiro}
\affiliation{
\institution{Sony}
}

\author{Chenfanfu Jiang}
\affiliation{
\institution{UCLA}
}

\makeatletter
\let\@authorsaddresses\@empty
\makeatother

\makeatletter
\def\@ACM@checkaffil{%
    \if@ACM@instpresent\else
    \ClassWarningNoLine{\@classname}{No institution present for an affiliation}%
    \fi
    \if@ACM@citypresent\else
    \ClassWarningNoLine{\@classname}{No city present for an affiliation}%
    \fi
    \if@ACM@countrypresent\else
    \ClassWarningNoLine{\@classname}{No country present for an affiliation}%
    \fi
}
\makeatother

\renewcommand{\shortauthors}{Z. Tao, Y. Jiang, Y. Chen, T. Xie, H. Wang, Y. Wu, Y. Yang, A. S. Kumar, K. Tashiro, C. Jiang}

\usepackage{algorithm}
\usepackage{amsmath}

\usepackage{algorithmicx}
\usepackage{algpseudocode}
\usepackage{multirow}
\usepackage{caption}
\usepackage{subcaption,bm}
\usepackage{ stmaryrd }
\usepackage{wrapfig}
\algdef{SE}[DOWHILE]{Do}{doWhile}{\algorithmicdo}[1]{\algorithmicwhile\ #1}
\AtBeginDocument{%
  \providecommand\BibTeX{{%
    \normalfont B\kern-0.5em{\scshape i\kern-0.25em b}\kern-0.8em\TeX}}}

\begin{document}

\begin{abstract}
Recent advances in garment pattern generation have shown promising progress. However, existing feed-forward methods struggle with diverse poses and viewpoints, while optimization-based approaches are computationally expensive and difficult to scale. This paper focuses on sewing pattern generation for garment modeling and fabrication applications that demand editable, separable, and simulation-ready garments. We propose \textbf{DressWild}, a novel feed-forward pipeline that reconstructs physics-consistent 2D sewing patterns and the corresponding 3D garments from a single in-the-wild image. Given an input image, our method leverages vision–language models (VLMs) to normalize pose variations at the image level, then extract pose-aware, 3D-informed garment features. These features are fused through a transformer-based encoder and subsequently used to predict sewing pattern parameters, which can be directly applied to physical simulation, texture synthesis, and multi-layer virtual try-on. Extensive experiments demonstrate that our approach robustly recovers diverse sewing patterns and the corresponding 3D garments from in-the-wild images without requiring multi-view inputs or iterative optimization, offering an efficient and scalable solution for realistic garment simulation and animation.
\end{abstract}

\ccsdesc[500]{Computing methodologies~Computer graphics}

\keywords{Sewing Pattern, Feed-forward Generation, In-the-wild, Vision-language Models}

\begin{teaserfigure}
  \centering
  \includegraphics[width=\textwidth]{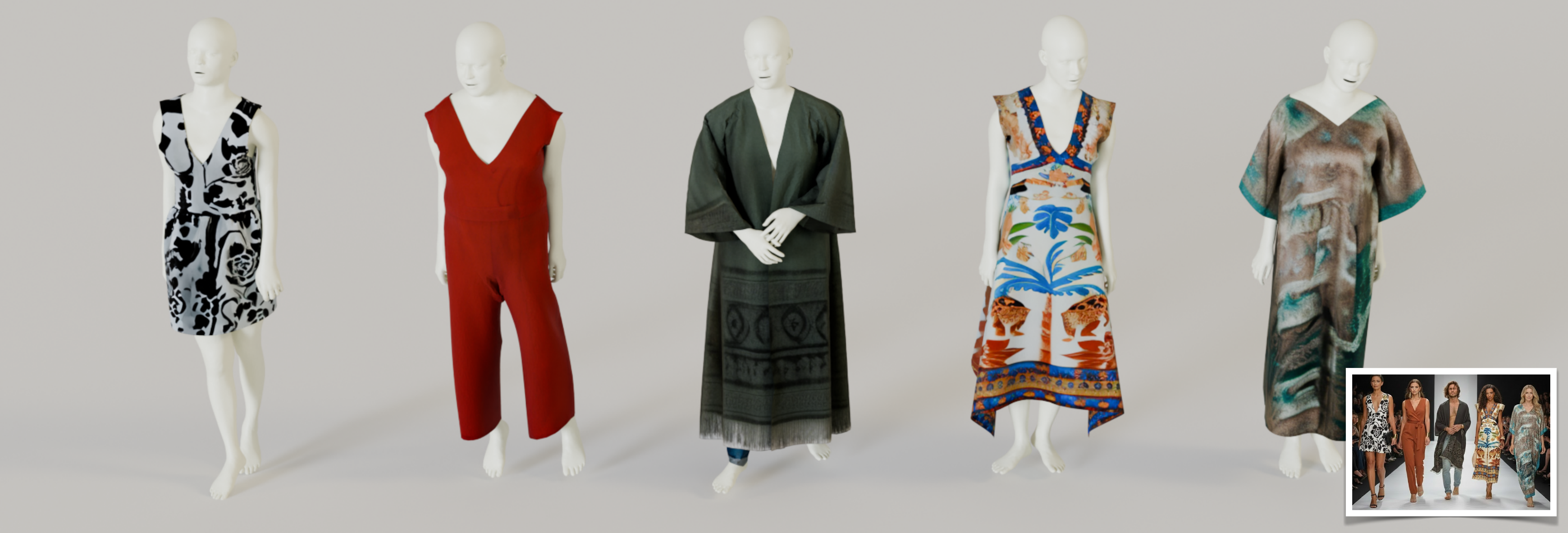}
  \caption{We propose \textbf{DressWild}, a feed-forward sewing-pattern generation pipeline that reconstructs simulation-ready, pose-agnostic 2D sewing patterns from a single in-the-wild image, while producing the corresponding pose-fitted 3D garment.}
  \label{fig:teaser}
\end{teaserfigure}

\maketitle

\section{Introduction}

Traditional 3D garment creation follows a multi-stage pipeline that progresses from concept design to 2D sewing pattern drafting and finally to 3D garment assembly through virtual sewing and garment simulation. While this workflow enables precise control and fabrication-ready outputs, it is inherently time-consuming and requires substantial domain expertise, making it inaccessible to non-experts. Recent advances in visual computing and artificial intelligence have significantly lowered the barrier to 3D garment creation by enabling direct generation from images \cite{bian2025chatgarment, li2025single}, text prompts \cite{li2025garmentdreamer, zhou2025design2garmentcode}, or point clouds \cite{korosteleva2022neuraltailor}, enabling efficient and scalable downstream virtual try-on and fabrication-aware garment design \cite{wu2025real}. However, most existing approaches focus solely on producing visually plausible 3D garment geometry, without recovering the underlying 2D sewing patterns. The absence of pattern-level representations limits editability, parametric control, and physical manufacturability, posing a fundamental challenge for downstream design iteration and real-world fabrication.

Existing approaches for sewing-pattern-based garment generation can be broadly categorized into data-driven feed-forward methods \cite{sewformer, liu2025multimodal, korosteleva2022neuraltailor} and optimization-based pipelines \cite{li2025dress, li2024diffavatar}. Data-driven methods leverage paired pattern–garment datasets to directly predict sewing patterns or 3D garments but their performance is tightly coupled to the training distribution, often restricting them to canonical body poses such as A- or T-poses and limiting their ability to generalize to diverse pose sewing pattern. Moreover, these approaches typically require multiple input images or controlled capture conditions, reducing their practicality in real-world scenarios. Optimization-based methods, on the other hand, explicitly enforce physical consistency between sewing patterns and 3D garments and can support multi-pose configurations; however, they rely on iterative simulation and gradient-based optimization, resulting in high computational cost and long runtimes. These limitations motivate the need for a more efficient framework that can recover sewing patterns and physically consistent 3D garments from an image.

To address these challenges, we introduce DressWild, a feed-forward sewing-pattern generation pipeline that reconstructs pose-agnostic 2D sewing patterns from a single in-the-wild image, while producing the corresponding pose-fitted 3D garment by leveraging VLM priors. Given an in-the-wild input image, we first leverage a VLM to generate an auxiliary standard T-pose 2D representation of the dressed person. We then perform garment segmentation on both the canonical T-pose image and the original input image, and extract reconstruction features from the segmented outputs to capture the garment’s structural and appearance cues while suppressing background clutter.
In parallel, we perform pose feature extraction on the input image to explicitly encode body articulation. These features are subsequently integrated through our feature fusion module with a hybrid attention strategy, which injects VLM priors while selectively attending to complementary structure and pose cues. We predict the 2D sewing pattern representation based on the fused features. To summarize, our contributions are:
\begin{itemize}
    \item We propose a feed-forward garment reconstruction pipeline that predicts diverse 2D sewing patterns and the corresponding physically consistent 3D garments, accurately fitted to people in arbitrary poses from a single in-the-wild image.
    \item We introduce a feature fusion and hybrid attention design that effectively incorporates VLM priors, enabling robust sewing pattern recovery and garment reconstruction under challenging multi-pose configurations.
    \item We conduct extensive experiments demonstrating the effectiveness and versatility of our approach, successfully reconstructing diverse sewing patterns and multi-pose 3D garments from in-the-wild images.
\end{itemize}

\begin{figure*}[t]
    \centering
    \includegraphics[width=0.98\linewidth]{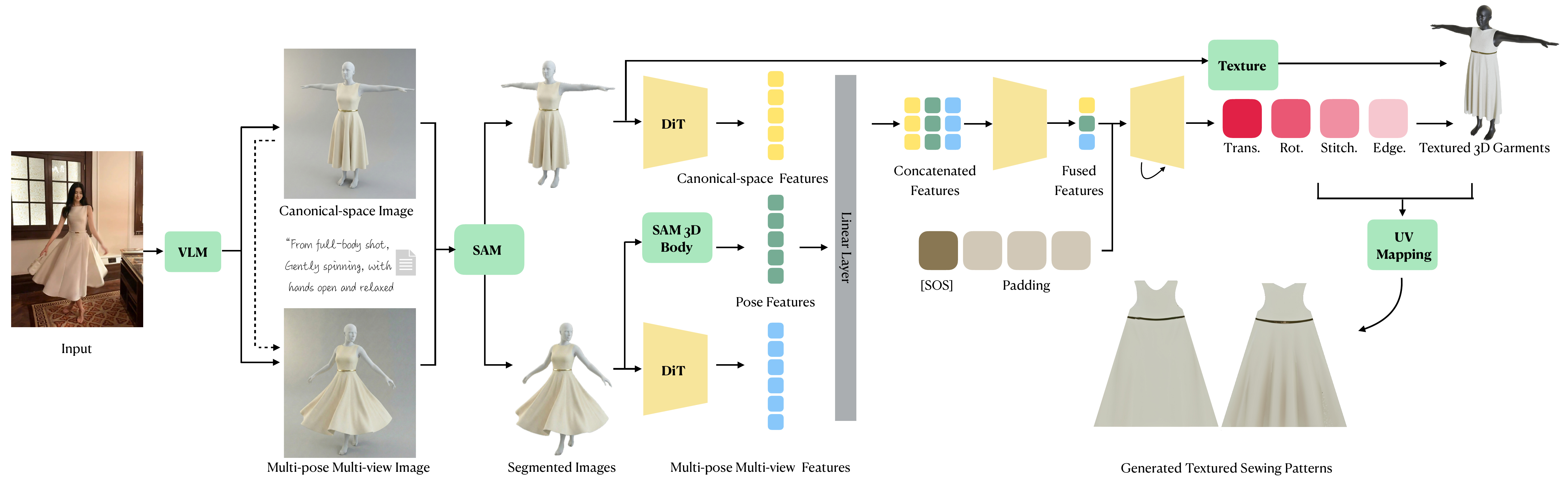}
    \caption{Overview of our pipeline. Given a single in-the-wild image, our method reconstructs simulation-ready sewing patterns and a corresponding 3D garment.}
    \label{fig:pipeline}
    \vspace{-10pt}
\end{figure*}

\section{Related Work}

\subsection{3D Reconstruction} 

With the emergence of powerful generative models, diffusions, transformers, autoregressive models and GANs have been widely adopted to synthesize 3D shapes from sparse conditioning signals such as images \cite{voleti2024sv3d, qian2023magic123, wang2025vggt, xiang2025structured}, videos \cite{cong2025videolifter}, text prompts \cite{chen2023fantasia3d, poole2022dreamfusion}, and sketches \cite{li2025meshpad}. Existing training strategies for 3D generative models generally fall into two categories: 3D supervision and 2D prior guidance. The former learns a conditional mapping from paired sparse inputs to 3D ground truth representations (e.g., meshes \cite{wang2018pixel2mesh}, voxels \cite{xie2019pix2vox}, point clouds \cite{wang2025vggt}, or implicit fields \cite{yu2021pixelnerf}), but typically relies on large-scale aligned 2D–3D datasets. The latter leverages multi-view rendering and differentiable rendering objectives derived from pretrained 2D priors, yet often requires additional constraints to ensure multi-view consistency \cite{huang2025mv, shi2023zero123++, xu2023dmv3d}. Despite strong performance, these data-driven approaches are still limited by data distributions, making reconstruction from in-the-wild data particularly challenging \cite{li2025dress}. To enable more diverse 3D generation without requiring paired data, score distillation sampling (SDS) \cite{poole2022dreamfusion} and variational score distillation (VSD) \cite{wang2023prolificdreamer} are introduced to optimize 3D representations under pretrained diffusion priors. However, such optimization-based methods require time-consuming iterative refinement for each instance, hindering scalability. Different from optimization-based approaches, we target an efficient and scalable feed-forward framework for garment generation from an in-the-wild image.

\vspace{-5px}
\subsection{Garment Modeling and Generation} 
A wide range of representations has been explored for modeling garment geometry, spanning implicit 3D fields, such as unsigned distance fields (UDF)~\cite{yu2024surf}, signed distance fields (SDF)~\cite{moon20223d}, neural radiance fields (NeRF)~\cite{}, and explicit 3D representations, including meshes, point clouds, and 3D Gaussian splats~\cite{wang2025semanticgarment, li2025garmentdreamer, sarafianos2025garment3dgen}, as well as 2D sewing pattern representations~\cite{sewformer, he2024dresscode}. Among these, many existing 3D garment generation methods adopt template-based pipelines to produce garment geometry. Specifically, BCNet~\cite{su2021bcnet}, DeepGarment~\cite{danvevrek2017deepgarment}, GarNet~\cite{gundogdu2019garnet}, and TailorNet~\cite{patel2020tailornet} employ feed-forward neural networks trained on synthetic datasets to predict template-based garment deformations, represented as per-vertex offsets or low-dimensional deformation latents, which are subsequently applied to predefined garment meshes. However, these methods typically require category-specific training, with each model assuming a fixed garment topology and being trained separately for individual garment types. Alternatively, \citet{liu2023modeling, li2025garmentdreamer, sarafianos2025garment3dgen, casado2022pergamo} adopt optimization-based pipelines that refine garment mesh vertices at test time using guidance from multi-view images, 3D Gaussian Splatting, latent CLIP supervision, or normal maps. Nevertheless, these approaches still rely on predefined garment mesh templates for initialization, which inherently constrains the diversity of garment shapes they can represent. To enable more diverse garment generation, recent works explore generalized generative models that synthesize garments from images or text instead of training separate models for each garment category and using predefines mesh templates.  SMPLicit~\cite{corona2021smplicit} introduces an MLP-based generative model that predicts unsigned distance fields for implicit cloth modeling in the SMPL body space. ClothWild~\cite{moon20223d} extends this formulation to in-the-wild images, jointly reconstructing human shape and clothing geometry from a single view. Surf-D~\cite{yu2024surf} further explores diffusion-based models to recover non-watertight garment geometry directly from images. These methods mainly target 3D garment geometry and might omit explicit 2D sewing patterns, thereby limiting their applicability to downstream simulation and animation.  Recent works DressCode~\cite{he2024dresscode} and SewFormer~\cite{sewformer}, explore transformer models to generate both 3D garments and corresponding 2D sewing patterns from text prompts and images, respectively. Our work also focuses on generating sewing patterns and their corresponding 3D garments.

\vspace{-5px}
\subsection{Sewing Pattern Generation}

The development of large-scale sewing pattern datasets, such as GarmentCodeData \cite{korosteleva2024garmentcodedata} and SewFactory \cite{liu2023towards}, has enabled data-driven approaches to sewing pattern generation. Building on these datasets, prior work explores generative models to infer structured sewing pattern representations in both implicit pixel space and explicit vector space, capturing 2D pattern geometry as well as the associated 3D transformation parameters required for garment assembly and simulation. For instance, learning-based approaches explore diffusions, transformers, autoregressive models \cite{he2024dresscode, chen2022structure, guo2025garmentx, liu2025multimodal, li2025garmagenet} to synthesize sewing patterns from images and text prompts. While these methods generate vectorized sewing pattern parameters, their discrete latent spaces can limit generalization to unseen garment styles. In contrast, Garment Image \cite{tatsukawa2025garmentimage} represents sewing pattern geometry, topology, and placement in a rasterized form, yielding a more continuous latent space and improved generalization. Similarly, ISP \cite{li2023isp}, GarmentRecovery \cite{li2024garment}, and DMap \cite{li2025single} adopt rasterized implicit sewing pattern representations to enable higher-quality garment reconstruction with improved efficiency. To further enable multimodal control for sewing pattern editing and garment generation from text, images, and sketches, recent methods such as AIpparel \cite{nakayama2025aipparel}, Design2GarmentCode \cite{zhou2025design2garmentcode}, and ChatGarment \cite{bian2025chatgarment} integrate large language models (LLMs), large multimodal models (LMMs), and vision–language models (VLMs), respectively, into the sewing pattern generation pipeline. However, these methods have difficulty generalizing to in-the-wild images, as reconstructed garments are often biased toward the training data distribution, resulting in misalignment with the input image and limiting the diversity and fidelity needed to capture real-world garment variations. To enable diverse garment generation, Dress-1-to-3 \cite{li2025dress} and DiffAvatar \cite{li2024diffavatar} adopt differentiable frameworks that optimize sewing pattern geometric and physical parameters, which are predicted by transformers or specified by users, respectively. These approaches enable high-fidelity reconstruction of out-of-distribution garment shapes. However, their reliance on differentiable simulation can lead to failure cases under large human motions or extreme poses and incurs high computational costs. To complement prior work on pattern geometry, DeepIron \cite{kwon2023deepiron} predicts unwarped sewing-pattern textures from single images given fixed sewing pattern shapes. In contrast, our work aims to efficiently generate diverse and pose-agnostic, simulation-ready textured sewing patterns from in-the-wild images via fused multi-pose feature representations.

\vspace{-10px}
\section{Method}

\subsection{Overview}

As shown in Fig.~\ref{fig:pipeline}, we present \textbf{DressWild}, a novel framework for diverse and pose-agnostic sewing pattern generation that supports a wide range of garment categories, including T-shirts, jackets, pants, skirts, jumpsuits, dresses, and other apparel types. Given an input in-the-wild image $I$, which may exhibit arbitrary pose and viewpoint, DressWild first explore a VLM like Nanobanana Pro~\cite{team2024gemini} to transform the input image $I$ into a canonical representation by synthesizing a garment image in a fixed T-pose and front-facing view $I_c$. This normalization is achieved using a pre-trained VLM, which provides strong priors for disentangling pose and viewpoint variations from garment appearance. By leveraging VLM priors, our framework aligns the normalized image $I_c$ to the pose and view distribution of the training dataset while preserving garment-specific visual details. Given $I_c$ and $I$, we extract image appearance features $f_i$, and canonical-space normalized image features $f_c$ using Hunyuan3D~\cite{lai2025hunyuan3d25highfidelity3d}. In parallel, we employ \textit{SAM3D-Body} \cite{yang2025sam3dbody}, a state-of-the-art human reconstruction method, to extract pose-aware features $f_p$ and recover a 3D human mesh $M_h$ from the input image $I$. These embeddings are fused to form a joint feature
\begin{equation}
f = \Phi(f_p, f_i, f_c),
\end{equation}
where $\Phi(\cdot)$ denotes a feature fusion module. $f$ is further processed by a feed-forward garment modeling network to predict pose- and view-invariant sewing pattern parameters.
 This design enables \textbf{DressWild} to leverage generative models trained on curated datasets while robustly generalizing to diverse, in-the-wild images without requiring multi-view or pose annotations. In the following sections, we elaborate on each component of the proposed pipeline.

\vspace{-5px}
\subsection{Data Processing}

\paragraph{Sewing Pattern Representation}

We follow the sewing pattern templates from \cite{korosteleva2022neuraltailor} and \cite{he2024dresscode}, covering diverse garment categories. We represent a garment sewing pattern as a set of 2D panels together with their 3D placement and stitching topology. Specifically, a pattern consists of $N_P$ panels $\{P_i\}_{i=1}^{N_P}$, where each panel $P_i$ is defined as a closed planar polygon with an ordered set of vertices $\{\bm{v}_{i,k}\}_{k=1}^{N_i} \subset \mathbb{R}^2$. Consecutive vertices form edges $\{E_{i,j}\}_{j=1}^{N_i}$ that describe the panel boundary. Each edge is defined by two endpoints and an optional curvature term. Straight edges are represented as line segments, while curved edges are modeled using quadratic Bézier curves. A curved edge is parameterized by its start point $\bm{v}_s$, end point $\bm{v}_e$, and a single control point $\bm{c} \in \mathbb{R}^2$ specified in relative coordinates. To enable panel meshing and subsequent physical simulation, we discretize each curved sewing
pattern edge by sampling points along its quadratic Bézier representation. Specifically, a point
on a curved edge is computed as $\bm{B}(t) = (1 - t)^2 \bm{v}_s + 2(1 - t)t \bm{c} + t^2 \bm{v}_e, t \in [0, 1]$, where uniformly sampling $t$ yields a set of boundary points that approximate the continuous
panel contour. The resulting discrete boundary enables robust panel triangulation for downstream cloth
simulation and stitching, while maintaining a compact and expressive representation of curved
garment contours. To place panels in 3D space, each panel $P_i$ is associated with a 6-DoF rigid transformation $(\bm{r}_i, \bm{T}_i)$, where $\bm{r}_i \in \mathbb{R}^3$ parameterizes the rotation and $\bm{T}_i \in \mathbb{R}^3$ denotes the translation, which defines its canonical 3D configuration. Stitching information is represented explicitly at the edge level using discrete stitching labels, where each label pairs two edges from different panels and encodes the sewing topology. Given an input image $I$, our network predicts the parameters of this structured representation, including panel vertex coordinates $\bm{v}_{i,k}$, edge curvature control points $\bm{c}_{i,j}$, panel-wise rigid transformations $(\bm{r}_i, \bm{T}_i)$, and stitching correspondences $\mathcal{S}$. By operating in this parametric sewing pattern space, the generated garments are pose- and view-invariant and directly compatible with physical simulation.

\vspace{-5px}
\paragraph{VLM-Guided Data Curation}
We curate a sewing pattern dataset for multi-pose and multi-view garment images, aiming to support learning based garment modeling under diverse body configurations and camera viewpoints. Our dataset is built upon the dataset of \cite{korosteleva2021generating}, which covers more than 20,000 garment design variations derived from 19 base garment types. The original dataset provides T pose images from only two viewpoints. To extend the input images to cover more diverse poses and viewpoints, we leverage a vision language model (VLM) to synthesize multi-view and multi-pose images from front view T pose inputs. Specifically, we define three configurable parameter sets, $n_v$ viewpoints, $n_p$ body poses, and $n_g$ gestures, which are randomly combined to construct prompt style inputs for image generation. Although our data curation framework supports additional augmentation dimensions such as background scenes, lighting conditions, and clothing material or pattern descriptors, we do not adopt these settings in our final dataset construction. In practice, introducing such variations increases synthesis instability and can lead to inconsistencies between the generated images and underlying garment geometry. From a learning perspective, these factors primarily affect background appearance or materials rather than garment structure, resulting in limited performance gains. Therefore, we restrict augmentation to pose, view, and gesture level variations.

More specifically, the viewpoint configuration includes short keyword prompts such as \textcolor{blue}{"low angle"}, \textcolor{blue}{"worm's-eye view"}, \textcolor{blue}{"Dutch angle"}, and \textcolor{blue}{"eye-level"}. The pose configuration is represented using concise descriptors including \textcolor{blue}{"standing"}, \textcolor{blue}{"arching back"}, \textcolor{blue}{"jumping"}, and \textcolor{blue}{"walking"}. The gesture configuration further augments motion diversity with prompt tokens such as \textcolor{blue}{"arms crossed"}, \textcolor{blue}{"hands on hips"}, and \textcolor{blue}{"wiping sweat from the forehead"}. This randomized combination of short, keyword-based prompts enables the generation of diverse and physically plausible variations while preserving garment structure and identity. In total, we define 10 distinct viewpoints, 21 body poses, and 21 gestures, whose randomized combinations are used as prompt tokens for multi-view and multi-pose image generation.

\begin{figure}[h]
    \centering
    \includegraphics[width=0.5\textwidth]{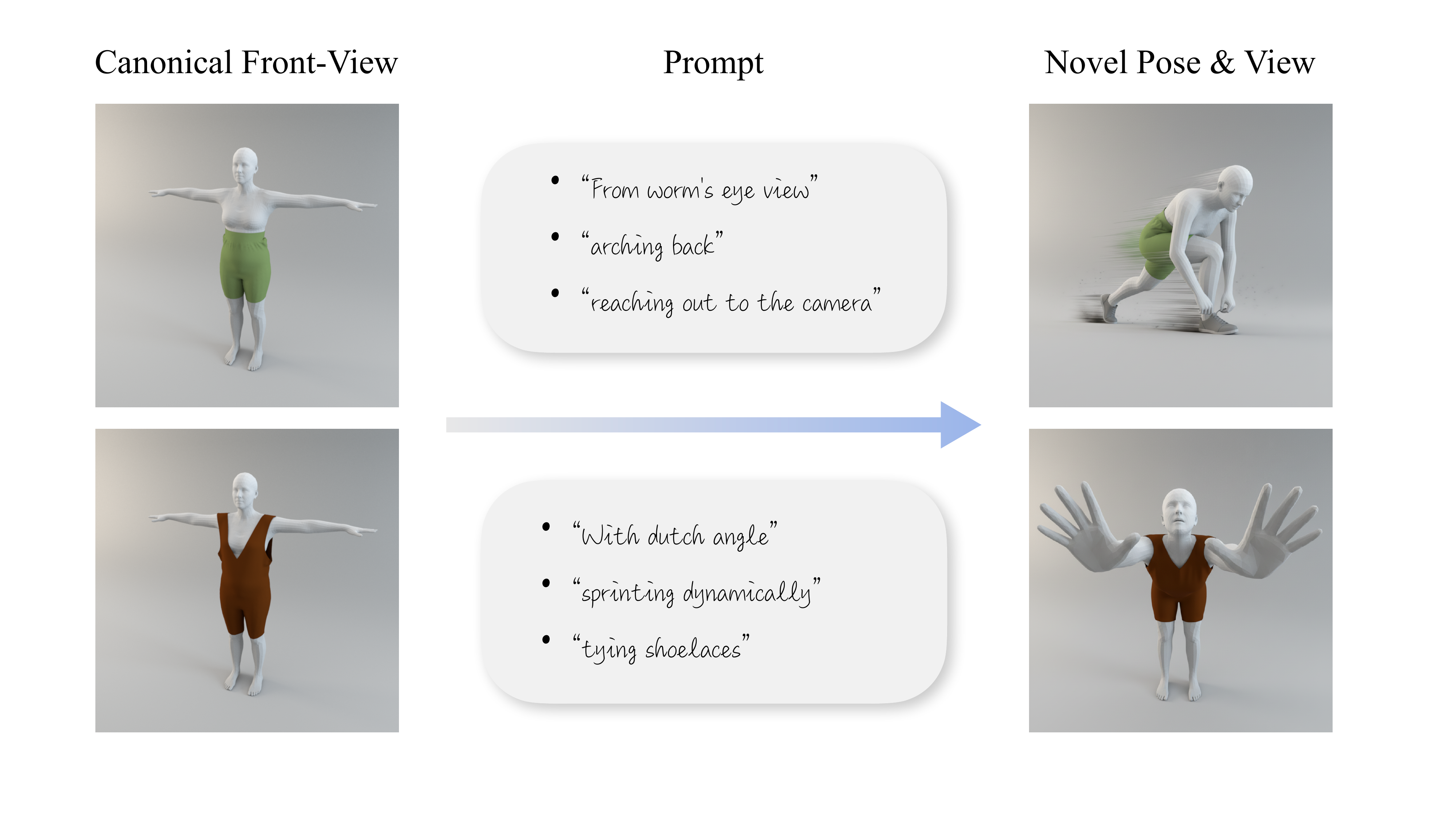}
    \vspace{-5px}
    \caption{\textbf{Data Curation and Augmentation.} We use VLM to generate novel pose and novel view images with the consistent garment.}
    \label{fig:data}
\end{figure}

\vspace{-15px}
\subsection{Feature Extraction}

Given an in-the-wild input image $I$, our goal is to extract pose-aware and pose-invariant garment features for robust sewing pattern prediction. To this end, we construct three complementary feature streams capturing image appearance, canonical garment structure, and human pose.

\paragraph{Canonicalization and Segmentation.}
We first leverage a VLM to normalize pose and viewpoint by synthesizing a canonical garment image $I_c$ in a fixed front-facing T-pose. This canonicalization step disentangles pose and view variations from garment appearance while preserving garment-specific visual details. We then apply HybridGL~\cite{liu2025hybrid} to segment both the original input image $I$ and the canonical image $I_c$, producing segmented garment images $I_{si}$ and $I_{sc}$, respectively.

\paragraph{3D Reconstruction Features.}
Based on the segmented images, we employ Hunyuan3D-2.0~\cite{hunyuan3d22025tencent} to extract 3D reconstruction features. Specifically, we utilize the intermediate latent representations produced after the Hunyuan3D-DiT sampling stage and before the VAE decoder, which capture rich 3D-aware garment geometry. From the segmented original image $I_{si}$, we extract multi-view multi-pose reconstruction features $f_i$, while from the canonical T-pose image $I_{sc}$, we obtain pose-normalized reconstruction features $f_c$. Both feature types are derived using the same reconstruction pipeline to ensure consistency.

\paragraph{Pose Features.}
In parallel, we apply SAM3D-Body~\cite{yang2025sam3dbody} to extract pose-aware embeddings $f_p$ by taking $I_{si}$ by the SAM3D-Body encoder. These features explicitly encode human pose information.

\vspace{-5px}
\subsection{Model}
\label{sec:model}

Our model consists of two stages including feature fusion and parameter decoding. Multimodal features are fused into a pose and view invariant representation $f$, which is then decoded to predict structured sewing pattern parameters $\Theta$ including edge, stitching, rotation, and translation information.

\paragraph{Feature Fusion Module.}
Since the three feature types originate from different encoders and exhibit heterogeneous dimensions and semantics, we first project each feature into a shared embedding space of dimension $d_w$ using separate linear transformations, $\tilde{f}_i = W_i f_i, \tilde{f}_c = W_c f_c, \tilde{f}_p = W_p f_p$
where $W_i$, $W_c$, and $W_p$ denote learnable projection matrices. This operation ensures that all features lie in the same feature space and can be jointly processed. We then concatenate the projected features along the sequence dimension, $\tilde{f} = [\tilde{f}_i;\tilde{f}_c;\tilde{f}_p],$
where the resulting sequence length equals $\lvert f_i\rvert + \lvert f_c\rvert + \lvert f_p\rvert$. The concatenated feature sequence is passed through a transformer encoder \cite{vaswani2017attention}, which performs self-attention across all tokens to capture cross modal interactions among image appearance, canonical garment structure, and pose information. This mechanism enables the model to adaptively aggregate complementary cues from different feature sources, producing a fused representation $f = \Phi(\tilde{f})$, where $\Phi(\cdot)$ denotes the feature fusion module.

\paragraph{Parameter Decoding.}
Given the fused feature representation $f$, we employ a decoder-based transformer~\cite{he2024dresscode} to autoregressively predict the sewing pattern parameters.

Each input token is formed by the sum of three embeddings: a positional embedding that identifies the panel associated with the token, a parameter type embedding that distinguishes edge geometry, rigid transformations, and stitching attributes, and a value embedding that encodes the quantized sewing pattern parameters.
Each output token represents a specific element of the sewing pattern parameterization, including panel vertex coordinates, edge curvature control points, 3D translation, rotation, and stitching correspondences. Through iterative decoding, the decoder attends to the fused feature representation and generates a sequence of latent tokens, which are subsequently decoded into the structured sewing pattern parameters.

\[
\Theta = \{\{\bm{v}_{i,k}\}, \{\bm{c}_{i,j}\}, \{(\bm{r}_i, \bm{T}_i)\}, \mathcal{S}\}.
\]

By conditioning the decoder on pose and view invariant fused features, the predicted sewing patterns are robust to pose and viewpoint variations and are directly compatible with downstream physical simulation and animation.

\paragraph{Training Objective.}
Our training objective consists of two components: a categorical cross entropy loss for discrete token prediction and a regression loss for continuous sewing pattern parameters. Our training objective is formulated as maximizing the joint conditional likelihood of discrete sewing pattern tokens and continuous geometric parameters. At each decoding step $t$, the decoder predicts a categorical distribution over discrete pattern parameter tokens, corresponding to the conditional likelihood
\[
\mathcal{L}_{\text{CE}}
= - \sum_{t=1}^{L} \log p(f_t \mid f_{<t},\, f;\theta),
\]
where $f_t$ denotes the ground truth token at step $t$, $f_{<t}$ represents all previously generated tokens, and $f$ is the fused conditioning feature. This formulation is equivalent to the standard token-wise cross entropy loss used in autoregressive sequence modeling. In addition to discrete token prediction, we model the decoded continuous sewing pattern parameters using a Gaussian likelihood. After mapping the predicted tokens back to their continuous parameter values, we assume an isotropic Gaussian distribution over the continuous parameters, whose negative log likelihood reduces to a mean squared error loss:
\[
\mathcal{L}_{\text{MSE}}
= - \sum_{t=1}^{L} \log p(\boldsymbol{\theta}_t \mid f_t,\, f;\theta)
\propto
\sum_{t=1}^{L} \lVert \hat{\boldsymbol{\theta}}_t - \boldsymbol{\theta}_t \rVert_2^2,
\]
where $\hat{\boldsymbol{\theta}}_t$ and $\boldsymbol{\theta}_t$ denote the predicted and ground truth continuous parameters at step $t$, respectively. This term encourages numerical accuracy of geometric quantities such as panel vertex coordinates and curvature control points. The final training objective is defined as a weighted sum of the two negative log likelihood terms:
\[
\mathcal{L}
= \mathcal{L}_{\text{CE}} + \lambda \mathcal{L}_{\text{MSE}},
\]
where $\lambda$ balances discrete token likelihood maximization and continuous parameter regression.

\section{Post Process}

\subsection{Texture Generation}

\paragraph{Texture Generation.}
The final step of \textbf{DressWild} focuses on generating high-fidelity textures for both sewing patterns, and their corresponding garment meshes, with the goal of supporting downstream garment fabrication. To this end, we introduce a dedicated texture generation module to reconstruct garment appearance. Unlike approaches that jointly generate geometry and texture, we explicitly decouple appearance synthesis from geometric reconstruction. Jointly predicting 2D sewing pattern textures and garment geometry is severely limited by the scarcity of paired supervision, making end-to-end learning difficult to scale. In addition, joint 3D texture and geometry generation from images is inherently ill-posed, as visual observations ambiguously entangle illumination, material appearance, and fine-scale geometric details, hindering reliable disentanglement of texture from shape.

Based on this design choice, we adopt a generative approach to first synthesize textures on the reconstructed 3D garment surface and subsequently transfer them to the sewing pattern domain. After obtaining sewing pattern parameters and generating a 3D garment mesh draped on the human body via physical simulation in Sec. \ref{sec:garmentsim}, we extract multi-view consistent garment textures using Hunyuan3D-Paint~\cite{hunyuan3d22025tencent}, which goes through image delighting, multi-view consistent image generation, super-resolution and baking. The generated textures are then projected onto the UV parameterization induced by the sewing patterns, ensuring seam consistency and direct applicability to pattern-based fabrication workflows. This separation allows textures to be consistently aligned across seams, robust to pose-induced deformations, and directly applicable to physical garment production.

\subsection{Garment Simulation} 
\label{sec:garmentsim}

We adopt the SMPL-X model~\cite{pavlakos2019expressive} as the articulated human body representation and initialize garment placement around a T-posed body following Dress123 \cite{li2025dress}, using Magicman \cite{he2025magicman} to provide a coarse initial fit. As this rough fitting often introduces interpenetrations, we first apply Position-Based Dynamics (PBD) to project the garment mesh outside the human body while enforcing high stretching and bending stiffness to preserve the original shape. To ensure proper layering for multi-garment cases, stitched vertices are used to identify connected components, which are sorted vertically and sequentially fitted from bottom to top using the Codimensional Incremental Potential Contact (CIPC) simulator~\cite{li2020codimensional}, allowing upper garments to naturally overlay lower ones. The resulting configuration serves as an initial feasible state for CIPC-based dynamic simulation, after which the human body is interpolated from the T-pose to the reconstructed pose and simulated as a moving boundary condition. To prevent slippage of lower garments during pose interpolation, we locally shrink the rest shape of triangles near the waist to induce sufficient friction; the garment rest pose otherwise remains unchanged.

\section{Experiments}

\subsection{Experiment Settings}
\label{sec:imp}
\paragraph{Implementation Details} We employ NanoBanana Pro~\cite{team2024gemini} as the VLM. For image feature extraction, we utilize the pre-trained Hunyuan3D-2~\cite{hunyuan3d22025tencent} model. We employ the pre-trained HybridGL~\cite{liu2025hybrid} model for referring segmentation, and SAM3D-Body Dino-v3~\cite{yang2025sam3dbody} is used for pose feature extraction. Regarding the model architecture, the decoder consists of 24 layers with 8 attention heads. We use the Adam optimizer with an initial learning rate set to $10^{-4}$. The model is trained for 10 epochs on a single NVIDIA H100 GPU.

\paragraph{Dataset} We utilize \cite{KorostelevaGarmentData} and our generated data to train and evaluate the models. The sewing patterns are categorized into 12 classes, which contain 25031 data points. We split the dataset into a training set, a validation set, and a testing set in a ratio of 7:2:1.

\paragraph{Baselines} We conduct both qualitative and quantitative comparisons with state-of-the-art sewing pattern generation methods, NeuralTailor and SewFormer, which generate sewing patterns from point clouds and single images, respectively. 

\subsection{Quantitative Comparison}
\label{sec:quan}

We conduct quantitative comparisons with the baseline methods on the same test set across multiple evaluation metrics, including panel accuracy, edge accuracy, shape reconstruction error, Filtered shape (F-shape) error, and Chamfer Distance (CD). NeuralTailor relies on garment point cloud inputs, while SewFormer and our method predict sewing patterns directly from single in-the-wild images. For fair comparison, we use Hunyuan3D to generate the mesh from the input image and sample point clouds as inputs to NeuralTailor. As shown in Table~1, NeuralTailor performs poorly on in-the-wild images, achieving only 25.99\% panel accuracy and 29.05\% edge accuracy, which reflects its limited robustness to pose and viewpoint variations. SewFormer improves structural prediction to some extent, reaching 28.81\% panel accuracy and 34.56\% edge accuracy, but still struggles to recover correct panel topology and geometry under arbitrary poses. In contrast, our method substantially outperforms both baselines, achieving 94.35\% panel accuracy and 85.41\% edge accuracy. Moreover, our approach yields significantly lower geometric errors, reducing the $\ell_2$ shape error from 23.65 for NeuralTailor and 22.94 for SewFormer to 6.22, and achieving the lowest Chamfer Distance of 0.01899. Sewformer performs slightly better on F-shape error, for this metric ignores the wrong panels. These results demonstrate that our method produces both structurally correct and geometrically accurate sewing patterns from challenging in-the-wild images with diverse poses and viewpoints.

\subsection{Qualitative Comparison}

\begin{figure*}[thbp]
    \centering
    \small
    \setlength{\tabcolsep}{2pt} 
    
    \begin{tabular}{cl}
        & \begin{tabular}{ccc}
            \makebox[0.3\textwidth]{Input} & \makebox[0.3\textwidth]{Input} & \makebox[0.3\textwidth]{Input}
          \end{tabular} \\
        
        \begin{tabular}{r}
            \rotatebox{90}{GT} \\[0.8cm]
            \rotatebox{90}{Ours} \\[1.2cm]
            \rotatebox{90}{NeuralTailor} \\[0.7cm]
            \rotatebox{90}{SewFormer}
        \end{tabular} & 
        \begin{minipage}{0.93\textwidth}
            \includegraphics[width=\linewidth]{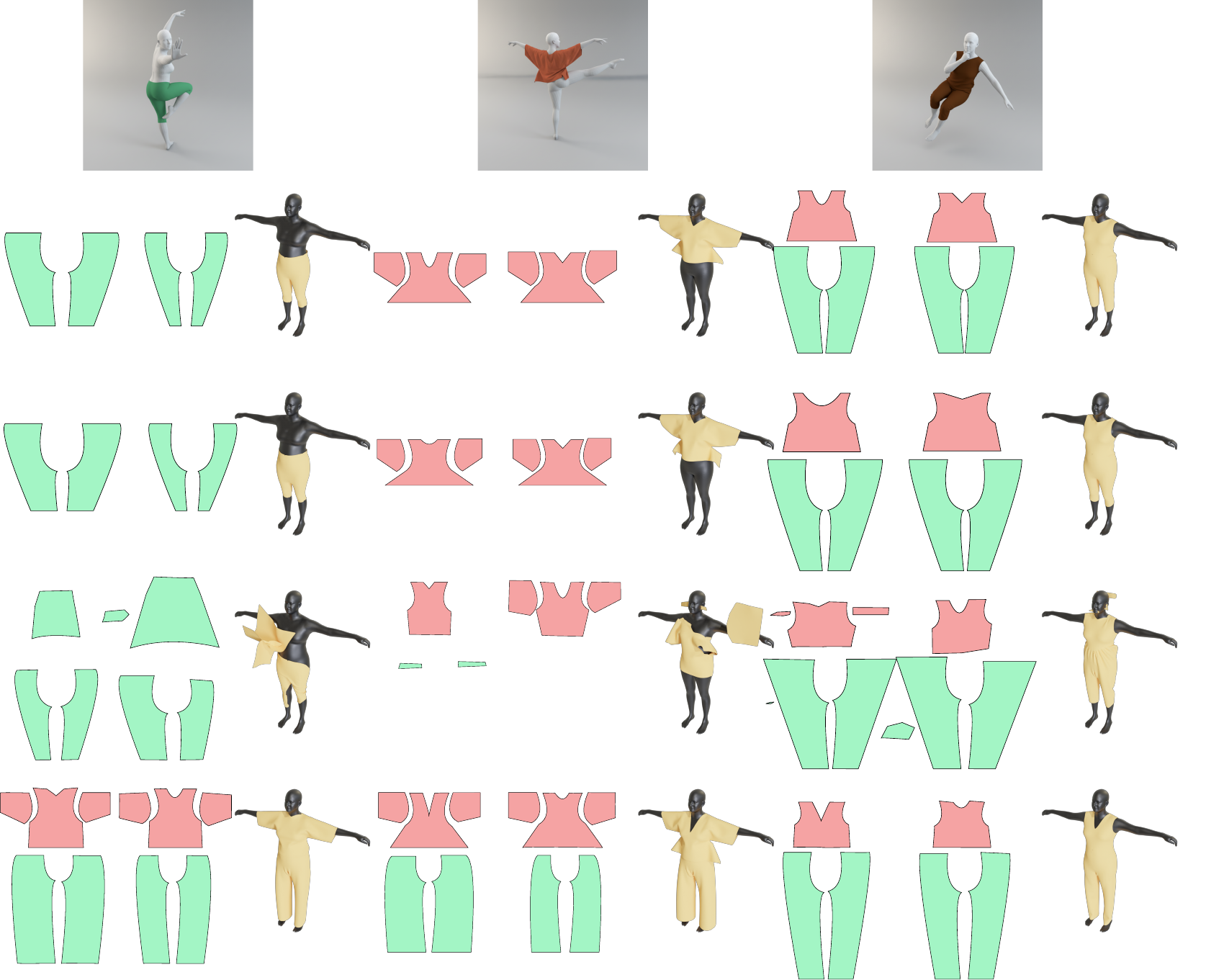}
        \end{minipage}
    \end{tabular}

    \caption{\textbf{Qualitative Comparison.} Our method produces high-quality 2D sewing patterns and corresponding 3D garments. Here we qualitatively compare our results with baseline approaches SewFormer~\cite{sewformer} and NeuralTailor~\cite{korosteleva2022neuraltailor}.}
    \label{fig:qualitative}
\end{figure*}

Fig.~\ref{fig:qualitative} presents qualitative comparisons with baseline methods on the same test set described in Sec.~\ref{sec:imp}, focusing on sewing pattern prediction from in-the-wild images with arbitrary poses and viewpoints. NeuralTailor, which relies on point cloud inputs, struggles to recover complete and structurally consistent pattern pieces, especially for garments with complex topology such as dresses and jackets. SewFormer shows reasonable performance under limited viewpoint variation, but its predictions often degrade for non-frontal or rotated poses, resulting in fragmented or misaligned panels. In contrast, our method consistently reconstructs coherent and well-proportioned sewing patterns across diverse in-the-wild conditions. By internally leveraging canonicalized representations together with original-view information, our approach better preserves garment topology, symmetry, and semantic part correspondence. These results demonstrate the robustness of our method to pose and viewpoint variations and its ability to recover simulation-ready sewing patterns from challenging in-the-wild images.

\begin{table}[h]
    \centering
    \caption{\textbf{Quantitative Comparison.} Our model outperforms the other methods on most of the metrics.}
    \scalebox{0.8}{
    \begin{tabular}{c|c|c|c|c|c}
        \toprule \toprule 
        Method & Panel Acc. $\uparrow$ & Edge Acc. $\uparrow$ & Shape $\mathcal{L}^2$ $\downarrow$ & F-Shape $\mathcal{L}^2$ $\downarrow$& CD $\downarrow$\\
        \midrule \midrule
        NeuralTailor & 25.99\% & 29.05\% & 23.65  & 7.30& 0.02837\\
        \midrule
        Sewformer & 28.81\% & 34.56\% & 22.94 & \textbf{4.53} &0.02797\\
        \midrule 
        \textbf{DressWild} & \textbf{94.35}\% &  \textbf{85.41}\% & \textbf{6.22} & 5.07 & \textbf{0.01899}\\
        
        \bottomrule \bottomrule
    \end{tabular}}
    \label{tab:quantitative}
\end{table}

Table.~\ref{tab:quantitative} presents our quantitative results. Panel Acc. and Edge Acc. denote the prediction accuracy for the number of panels and edges, respectively, while Shape $\mathcal{L}^2$ Error represents the $\mathcal{L}^2$ distance of the parameters. Our method outperforms Sewformer across most of the metrics.

\subsection{Ablation Study}

We conduct an ablation study of the key components of our framework including additional visual features (canonical-space reconstruction features, pose features), and the feature fusion module, using the same garment images as in Sec. \ref{sec:imp}. This study assesses how each proposed component contributes to the final sewing pattern reconstruction quality.

\begin{figure*}[t]
    \centering
    \includegraphics[width=0.95\textwidth]{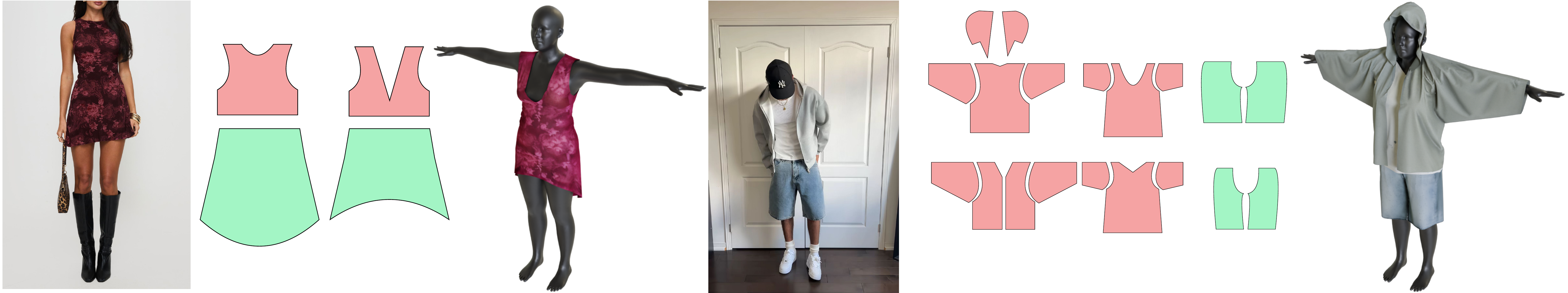}
    
    \vspace{1mm}
    \small
    \makebox[0.15\textwidth]{Input} \hfill 
    \makebox[0.15\textwidth]{Sewing patterns} \hfill 
    \makebox[0.15\textwidth]{Garments} \hfill 
    \makebox[0.15\textwidth]{Input} \hfill 
    \makebox[0.15\textwidth]{Sewing patterns} \hfill 
    \makebox[0.15\textwidth]{Garments}
    
    \caption{\textbf{Results from in-the-wild image with multi-layer garments.} Given an in-the-wild image as input, our DressWild method can generate multi-layer pose-agnostic and simulation-ready sewing patterns.}
    \label{fig:wild}
\end{figure*}

\begin{table}[h]
    \centering
    \caption{\textbf{Ablation of canonical-space reconstruction features.} The performance of sewing pattern reconstruction is significantly improved with the inclusion of this feature.}
    \scalebox{0.8}{
    \begin{tabular}{c|c|c|c}
        \toprule \toprule 
        Method & Panel Acc. $\uparrow$ & Edge Acc. $\uparrow$ & Shape $\mathcal{L}^2$ $\downarrow$\\
        \midrule
        w/o front features & 91.18\% & 78.58\% & 12.11 \\
        \midrule
        w/ front features &  94.35\% &  85.41\% & 6.22   \\
        \bottomrule \bottomrule
    \end{tabular}}
    \label{tab:ablation_feature}
    \vspace{-10pt}
\end{table}

\begin{figure}[h]
    \centering
    \vspace{2mm}
    \begin{overpic}[width=0.47\textwidth]{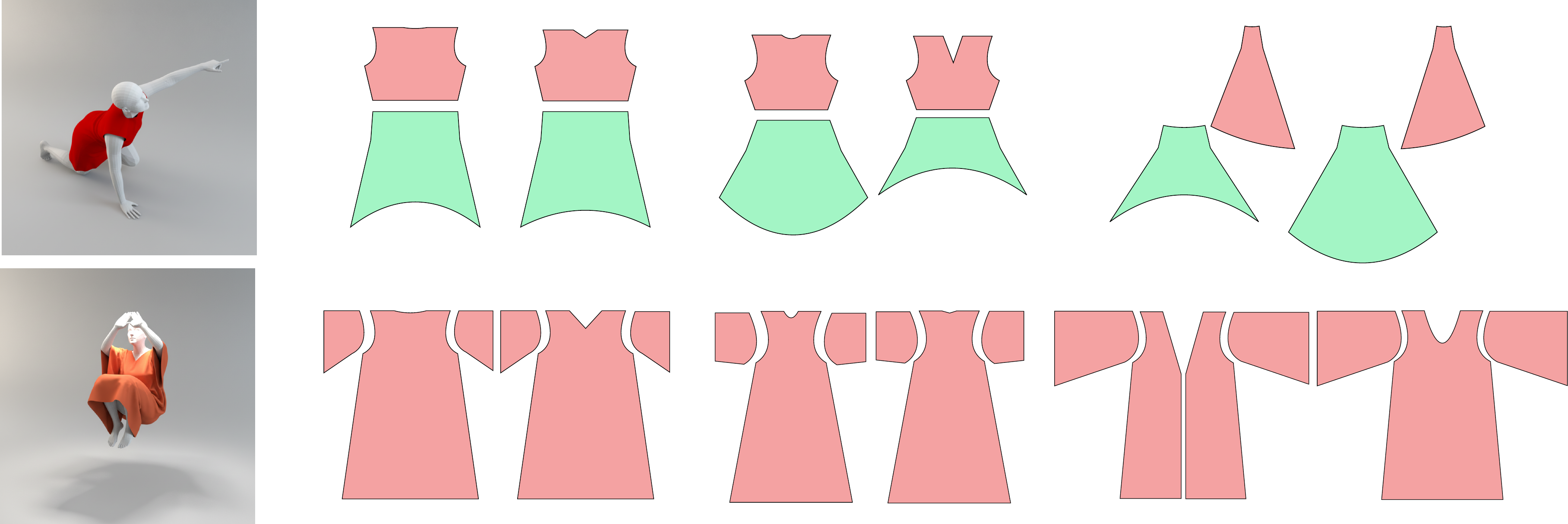}
        \put(5,35){\small Input}
        \put(30,35){\small GT}
        \put(48,35){\small w/ Pose Feat.}
        \put(73,35){\small w/o Pose Feat.}
    \end{overpic}
    \caption{\textbf{Ablation of pose features.} The model identifies the garment better with the assistance of the pose feature when the human pose is challenging.}
    \label{fig:ablation_feature}
    \vspace{-20pt}
\end{figure}

\paragraph{Feature Extraction.} We first evaluate the effect of different feature choices in our framework. As shown in Tab.~\ref{tab:ablation_feature}, incorporating canonical-space, front-view T-pose reconstruction features $f_c$ consistently improves both panel and edge accuracy, while reducing the shape error, indicating that front features provide critical geometric cues for recovering more accurate panel boundaries and overall sewing pattern shapes. Regarding pose features $f_p$ (Fig.~\ref{fig:ablation_feature}), predictions that incorporate pose information remain much closer to the ground truth in both panel shape and overall layout, producing more plausible boundaries. By providing explicit pose information, the model is able to disentangle pose-induced garment deformation from the underlying, pose-invariant pattern geometry. In contrast, removing pose features causes the network to misinterpret draped appearances under non-neutral poses as intrinsic pattern geometry, leading to distorted or warped panels and incorrect proportions.

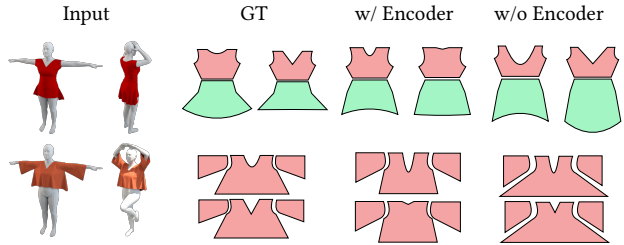
\begin{figure}[h]
    \centering
    \vspace{2mm}
    \hspace{-5mm}
    \begin{overpic}[width=0.47\textwidth]{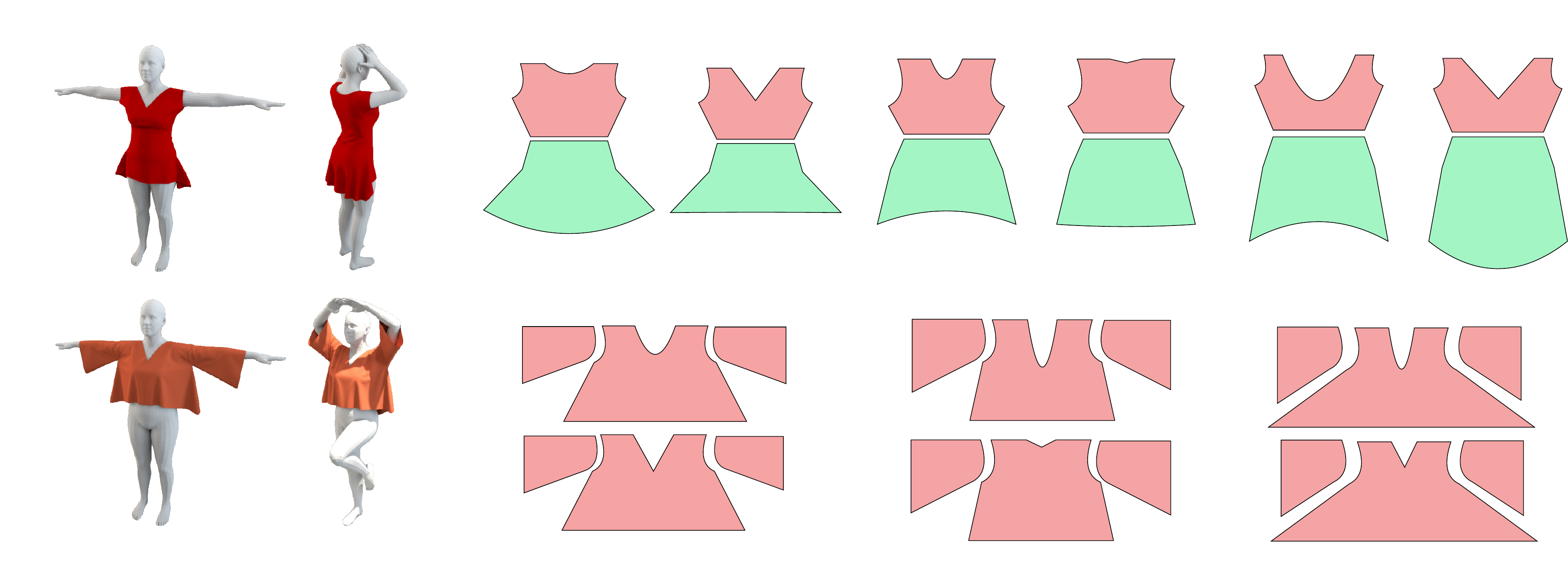}
        \put(12,37){\small Input}
        \put(40,37){\small GT}
        \put(58,37){\small w/ Encoder}
        \put(80,37){\small w/o Encoder}
    \end{overpic}
    \caption{\textbf{Ablation of feature fusion.} With feature fusion design, the results are more consistent with GT.}
    \label{fig:ablation_fusion}
    \vspace{-20pt}
\end{figure}

\paragraph{Feature Fusion Encoder.}
Since the features $f_i$, $f_c$, and $f_p$ originate from different encoders, we project them into a shared embedding space and fuse them with a transformer encoder using self-attention, producing a unified representation $f$. As shown in Fig.~\ref{fig:ablation_fusion}, models equipped with the feature fusion module produce sewing pattern predictions that closely match the ground truth in both panel shape and overall layout. By projecting $f_i$, $f_c$, and $f_p$ into the shared embedding space, the model effectively captures cross-feature information.  In contrast, removing the feature fusion module significantly degrades reconstruction quality, leading to incorrect proportions and distorted silhouettes, such as deformed collar shapes and inconsistent panel geometry. This degradation indicates that without explicit self-attention–based fusion, the model struggles to coherently aggregate complementary cues from different feature sources.

\section{Application}

\paragraph{Sewing Pattern Generation from In-the-wild Images}

As illustrated in Fig.~\ref{fig:wild}, our method is capable of directly processing in-the-wild images exhibiting diverse poses, viewpoints. Given an in-the-wild input image, we apply \textbf{DressWild} to robustly generate pose-agnostic and simulation-ready sewing patterns.

\paragraph{Multi-layer Garment Generation}

For garments composed of multiple layers, we follow a similar strategy by leveraging the VLM to decompose the input into individual garment layer image. Specifically, given an in-the-wild image containing layered apparel (e.g., a jacket worn over a shirt), we utilize the VLM to identify and extract each single-layer garment while preserving their relative ordering and semantic consistency. Each extracted garment layer image is then processed independently using \textbf{DressWild} to predict its corresponding pose-agnostic and simulation-ready sewing pattern as indicated in Fig.~\ref{fig:wild}. This design allows our framework to naturally extend from single-layer to multi-layer garment scenarios without requiring explicit layer annotations or modifications to the core model.

\vspace{-5px}
\section{Conclusion}

In this paper, we present DressWild, a novel feed-forward framework for reconstructing pose-agnostic, simulation-ready 2D sewing patterns and corresponding 3D garments from a single in-the-wild image. By leveraging VLMs to generate canonical garment representations and integrating pose-aware, canonical-space, and multi-view features via a hybrid mechanism, our approach effectively disentangles garment geometry from pose and viewpoint variations. Extensive experiments demonstrate that DressWild outperforms existing state-of-the-art methods in both quantitative metrics and qualitative comparison, particularly on challenging in-the-wild data with diverse poses and viewpoints. Furthermore, our pipeline supports multi-layer garment generation and produces physically plausible outputs compatible with downstream simulation and texture synthesis workflows, offering a scalable and efficient solution for realistic garment modeling and animation applications.

\bibliographystyle{ACM-Reference-Format}
\bibliography{main}

@String{Computer = "{IEEE} Computer" }

@String{Springer = "Springer-Verlag" }

@article{sewformer,
  title={Towards garment sewing pattern reconstruction from a single image},
  author={Liu, Lijuan and Xu, Xiangyu and Lin, Zhijie and Liang, Jiabin and Yan, Shuicheng},
  journal={ACM Transactions on Graphics (TOG)},
  volume={42},
  number={6},
  pages={1--15},
  year={2023},
  publisher={ACM New York, NY, USA}
}

@inproceedings{li2025garmentdreamer,
  title={Garmentdreamer: 3dgs guided garment synthesis with diverse geometry and texture details},
  author={Li, Boqian and Li, Xuan and Jiang, Ying and Xie, Tianyi and Gao, Feng and Wang, Huamin and Yang, Yin and Jiang, Chenfanfu},
  booktitle={2025 International Conference on 3D Vision (3DV)},
  pages={1416--1426},
  year={2025},
  organization={IEEE}
}

@inproceedings{bian2025chatgarment,
  title={Chatgarment: Garment estimation, generation and editing via large language models},
  author={Bian, Siyuan and Xu, Chenghao and Xiu, Yuliang and Grigorev, Artur and Liu, Zhen and Lu, Cewu and Black, Michael J and Feng, Yao},
  booktitle={Proceedings of the Computer Vision and Pattern Recognition Conference},
  pages={2924--2934},
  year={2025}
}

@inproceedings{casado2022pergamo,
  title={Pergamo: Personalized 3d garments from monocular video},
  author={Casado-Elvira, Andr{\'e}s and Trinidad, Marc Comino and Casas, Dan},
  booktitle={Computer Graphics Forum},
  volume={41},
  number={8},
  pages={293--304},
  year={2022},
  organization={Wiley Online Library}
}

@inproceedings{danvevrek2017deepgarment,
  title={Deepgarment: 3d garment shape estimation from a single image},
  author={Dan{\v{e}}{\v{r}}ek, R and Dibra, Endri and {\"O}ztireli, Cengiz and Ziegler, Remo and Gross, Markus},
  booktitle={Computer Graphics Forum},
  volume={36},
  number={2},
  pages={269--280},
  year={2017},
  organization={Wiley Online Library}
}

@inproceedings{su2021bcnet,
  title={Bcnet: Searching for network width with bilaterally coupled network},
  author={Su, Xiu and You, Shan and Wang, Fei and Qian, Chen and Zhang, Changshui and Xu, Chang},
  booktitle={Proceedings of the IEEE/CVF Conference on Computer Vision and Pattern Recognition},
  pages={2175--2184},
  year={2021}
}

@inproceedings{corona2021smplicit,
  title={Smplicit: Topology-aware generative model for clothed people},
  author={Corona, Enric and Pumarola, Albert and Alenya, Guillem and Pons-Moll, Gerard and Moreno-Noguer, Francesc},
  booktitle={Proceedings of the IEEE/CVF conference on computer vision and pattern recognition},
  pages={11875--11885},
  year={2021}
}

@inproceedings{gundogdu2019garnet,
  title={Garnet: A two-stream network for fast and accurate 3d cloth draping},
  author={Gundogdu, Erhan and Constantin, Victor and Seifoddini, Amrollah and Dang, Minh and Salzmann, Mathieu and Fua, Pascal},
  booktitle={Proceedings of the IEEE/CVF international conference on computer vision},
  pages={8739--8748},
  year={2019}
}

@article{li2020codimensional,
  title={Codimensional incremental potential contact},
  author={Li, Minchen and Kaufman, Danny M and Jiang, Chenfanfu},
  journal={arXiv preprint arXiv:2012.04457},
  year={2020}
}

@inproceedings{he2025magicman,
  title={Magicman: Generative novel view synthesis of humans with 3d-aware diffusion and iterative refinement},
  author={He, Xu and Wu, Zhiyong and Li, Xiaoyu and Kang, Di and Zhang, Chaopeng and Ye, Jiangnan and Chen, Liyang and Gao, Xiangjun and Zhang, Han and Zhuang, Haolin},
  booktitle={Proceedings of the AAAI Conference on Artificial Intelligence},
  volume={39},
  number={3},
  pages={3437--3445},
  year={2025}
}

@inproceedings{pavlakos2019expressive,
  title={Expressive body capture: 3d hands, face, and body from a single image},
  author={Pavlakos, Georgios and Choutas, Vasileios and Ghorbani, Nima and Bolkart, Timo and Osman, Ahmed AA and Tzionas, Dimitrios and Black, Michael J},
  booktitle={Proceedings of the IEEE/CVF conference on computer vision and pattern recognition},
  pages={10975--10985},
  year={2019}
}

@inproceedings{patel2020tailornet,
  title={Tailornet: Predicting clothing in 3d as a function of human pose, shape and garment style},
  author={Patel, Chaitanya and Liao, Zhouyingcheng and Pons-Moll, Gerard},
  booktitle={Proceedings of the IEEE/CVF conference on computer vision and pattern recognition},
  pages={7365--7375},
  year={2020}
}

@article{vaswani2017attention,
  title={Attention is all you need},
  author={Vaswani, Ashish and Shazeer, Noam and Parmar, Niki and Uszkoreit, Jakob and Jones, Llion and Gomez, Aidan N and Kaiser, {\L}ukasz and Polosukhin, Illia},
  journal={Advances in neural information processing systems},
  volume={30},
  year={2017}
}

@inproceedings{sarafianos2025garment3dgen,
  title={Garment3dgen: 3d garment stylization and texture generation},
  author={Sarafianos, Nikolaos and Stuyck, Tuur and Xiang, Xiaoyu and Li, Yilei and Popovic, Jovan and Ranjan, Rakesh},
  booktitle={2025 International Conference on 3D Vision (3DV)},
  pages={1382--1393},
  year={2025},
  organization={IEEE}
}

@inproceedings{zhou2025design2garmentcode,
  title={Design2GarmentCode: Turning Design Concepts to Tangible Garments Through Program Synthesis},
  author={Zhou, Feng and Liu, Ruiyang and Liu, Chen and He, Gaofeng and Li, Yong-Lu and Jin, Xiaogang and Wang, Huamin},
  booktitle={Proceedings of the Computer Vision and Pattern Recognition Conference},
  pages={23712--23722},
  year={2025}
}

@article{korosteleva2022neuraltailor,
  title={Neuraltailor: Reconstructing sewing pattern structures from 3d point clouds of garments},
  author={Korosteleva, Maria and Lee, Sung-Hee},
  journal={ACM Transactions on Graphics (TOG)},
  volume={41},
  number={4},
  pages={1--16},
  year={2022},
  publisher={ACM New York, NY, USA}
}

@article{li2025garmagenet,
  title={GarmageNet: A Multimodal Generative Framework for Sewing Pattern Design and Generic Garment Modeling},
  author={Li, Siran and Liu, Ruiyang and Liu, Chen and Wang, Zhendong and He, Gaofeng and Li, Yong-Lu and Jin, Xiaogang and Wang, Huamin},
  journal={ACM Transactions on Graphics (TOG)},
  volume={44},
  number={6},
  pages={1--23},
  year={2025},
  publisher={ACM New York, NY, USA}
}

@inproceedings{wang2025semanticgarment,
  title={SemanticGarment: Semantic-Controlled Generation and Editing of 3D Gaussian Garments},
  author={Wang, Ruiyan and Cheng, Zhengxue and Lin, Zonghao and Ling, Jun and Liu, Yuzhou and An, Yanru and Xie, Rong and Song, Li},
  booktitle={Proceedings of the 33rd ACM International Conference on Multimedia},
  pages={9793--9802},
  year={2025}
}

@article{cong2025videolifter,
  title={Videolifter: Lifting videos to 3d with fast hierarchical stereo alignment},
  author={Cong, Wenyan and Zhu, Hanqing and Wang, Kevin and Lei, Jiahui and Stearns, Colton and Cai, Yuanhao and Wang, Dilin and Ranjan, Rakesh and Feiszli, Matt and Guibas, Leonidas and others},
  journal={arXiv preprint arXiv:2501.01949},
  year={2025}
}

@article{shi2023zero123++,
  title={Zero123++: a single image to consistent multi-view diffusion base model},
  author={Shi, Ruoxi and Chen, Hansheng and Zhang, Zhuoyang and Liu, Minghua and Xu, Chao and Wei, Xinyue and Chen, Linghao and Zeng, Chong and Su, Hao},
  journal={arXiv preprint arXiv:2310.15110},
  year={2023}
}

@inproceedings{huang2025mv,
  title={Mv-adapter: Multi-view consistent image generation made easy},
  author={Huang, Zehuan and Guo, Yuan-Chen and Wang, Haoran and Yi, Ran and Ma, Lizhuang and Cao, Yan-Pei and Sheng, Lu},
  booktitle={Proceedings of the IEEE/CVF International Conference on Computer Vision},
  pages={16377--16387},
  year={2025}
}

@article{poole2022dreamfusion,
  title={Dreamfusion: Text-to-3d using 2d diffusion},
  author={Poole, Ben and Jain, Ajay and Barron, Jonathan T and Mildenhall, Ben},
  journal={arXiv preprint arXiv:2209.14988},
  year={2022}
}

@inproceedings{yu2021pixelnerf,
  title={pixelnerf: Neural radiance fields from one or few images},
  author={Yu, Alex and Ye, Vickie and Tancik, Matthew and Kanazawa, Angjoo},
  booktitle={Proceedings of the IEEE/CVF conference on computer vision and pattern recognition},
  pages={4578--4587},
  year={2021}
}

@inproceedings{wang2025vggt,
  title={Vggt: Visual geometry grounded transformer},
  author={Wang, Jianyuan and Chen, Minghao and Karaev, Nikita and Vedaldi, Andrea and Rupprecht, Christian and Novotny, David},
  booktitle={Proceedings of the Computer Vision and Pattern Recognition Conference},
  pages={5294--5306},
  year={2025}
}

@inproceedings{xiang2025structured,
  title={Structured 3d latents for scalable and versatile 3d generation},
  author={Xiang, Jianfeng and Lv, Zelong and Xu, Sicheng and Deng, Yu and Wang, Ruicheng and Zhang, Bowen and Chen, Dong and Tong, Xin and Yang, Jiaolong},
  booktitle={Proceedings of the Computer Vision and Pattern Recognition Conference},
  pages={21469--21480},
  year={2025}
}

@inproceedings{chen2023fantasia3d,
  title={Fantasia3d: Disentangling geometry and appearance for high-quality text-to-3d content creation},
  author={Chen, Rui and Chen, Yongwei and Jiao, Ningxin and Jia, Kui},
  booktitle={Proceedings of the IEEE/CVF international conference on computer vision},
  pages={22246--22256},
  year={2023}
}

@article{qian2023magic123,
  title={Magic123: One image to high-quality 3d object generation using both 2d and 3d diffusion priors},
  author={Qian, Guocheng and Mai, Jinjie and Hamdi, Abdullah and Ren, Jian and Siarohin, Aliaksandr and Li, Bing and Lee, Hsin-Ying and Skorokhodov, Ivan and Wonka, Peter and Tulyakov, Sergey and others},
  journal={arXiv preprint arXiv:2306.17843},
  year={2023}
}

@inproceedings{voleti2024sv3d,
  title={Sv3d: Novel multi-view synthesis and 3d generation from a single image using latent video diffusion},
  author={Voleti, Vikram and Yao, Chun-Han and Boss, Mark and Letts, Adam and Pankratz, David and Tochilkin, Dmitry and Laforte, Christian and Rombach, Robin and Jampani, Varun},
  booktitle={European Conference on Computer Vision},
  pages={439--457},
  year={2024},
  organization={Springer}
}

@inproceedings{yu2024surf,
  title={Surf-D: Generating High-Quality Surfaces of Arbitrary Topologies Using Diffusion Models},
  author={Yu, Zhengming and Dou, Zhiyang and Long, Xiaoxiao and Lin, Cheng and Li, Zekun and Liu, Yuan and M{\"u}ller, Norman and Komura, Taku and Habermann, Marc and Theobalt, Christian and others},
  booktitle={European Conference on Computer Vision},
  pages={419--438},
  year={2024},
  organization={Springer}
}

@article{xu2023dmv3d,
  title={Dmv3d: Denoising multi-view diffusion using 3d large reconstruction model},
  author={Xu, Yinghao and Tan, Hao and Luan, Fujun and Bi, Sai and Wang, Peng and Li, Jiahao and Shi, Zifan and Sunkavalli, Kalyan and Wetzstein, Gordon and Xu, Zexiang and others},
  journal={arXiv preprint arXiv:2311.09217},
  year={2023}
}

@inproceedings{tatsukawa2025garmentimage,
  title={GarmentImage: Raster encoding of garment sewing patterns with diverse topologies},
  author={Tatsukawa, Yuki and Qi, Anran and Shen, I-Chao and Igarashi, Takeo},
  booktitle={Proceedings of the Special Interest Group on Computer Graphics and Interactive Techniques Conference Conference Papers},
  pages={1--11},
  year={2025}
}

@article{chen2022structure,
  title={Structure-preserving 3d garment modeling with neural sewing machines},
  author={Chen, Xipeng and Wang, Guangrun and Zhu, Dizhong and Liang, Xiaodan and Torr, Philip and Lin, Liang},
  journal={Advances in Neural Information Processing Systems},
  volume={35},
  pages={15147--15159},
  year={2022}
}

@article{li2025dress,
  title={Dress-1-to-3: Single Image to Simulation-Ready 3D Outfit with Diffusion Prior and Differentiable Physics},
  author={Li, Xuan and Yu, Chang and Du, Wenxin and Jiang, Ying and Xie, Tianyi and Chen, Yunuo and Yang, Yin and Jiang, Chenfanfu},
  journal={ACM Transactions on Graphics (TOG)},
  volume={44},
  number={4},
  pages={1--16},
  year={2025},
  publisher={ACM New York, NY, USA}
}

@inproceedings{wang2018pixel2mesh,
  title={Pixel2mesh: Generating 3d mesh models from single rgb images},
  author={Wang, Nanyang and Zhang, Yinda and Li, Zhuwen and Fu, Yanwei and Liu, Wei and Jiang, Yu-Gang},
  booktitle={Proceedings of the European conference on computer vision (ECCV)},
  pages={52--67},
  year={2018}
}

@inproceedings{moon20223d,
  title={3d clothed human reconstruction in the wild},
  author={Moon, Gyeongsik and Nam, Hyeongjin and Shiratori, Takaaki and Lee, Kyoung Mu},
  booktitle={European conference on computer vision},
  pages={184--200},
  year={2022},
  organization={Springer}
}

@inproceedings{xie2019pix2vox,
  title={Pix2vox: Context-aware 3d reconstruction from single and multi-view images},
  author={Xie, Haozhe and Yao, Hongxun and Sun, Xiaoshuai and Zhou, Shangchen and Zhang, Shengping},
  booktitle={Proceedings of the IEEE/CVF international conference on computer vision},
  pages={2690--2698},
  year={2019}
}

@article{he2024dresscode,
  title={Dresscode: Autoregressively sewing and generating garments from text guidance},
  author={He, Kai and Yao, Kaixin and Zhang, Qixuan and Yu, Jingyi and Liu, Lingjie and Xu, Lan},
  journal={ACM Transactions on Graphics (TOG)},
  volume={43},
  number={4},
  pages={1--13},
  year={2024},
  publisher={ACM New York, NY, USA}
}

@article{guo2025garmentx,
  title={GarmentX: Autoregressive Parametric Representations for High-Fidelity 3D Garment Generation},
  author={Guo, Jingfeng and Chen, Jinnan and Chen, Weikai and Sun, Zhenyu and Li, Lanjiong and Zhao, Baozhu and Zhu, Lingting and Wang, Xin and Liu, Qi},
  journal={arXiv preprint arXiv:2504.20409},
  year={2025}
}

@article{li2023isp,
  title={Isp: Multi-layered garment draping with implicit sewing patterns},
  author={Li, Ren and Guillard, Beno{\^\i}t and Fua, Pascal},
  journal={Advances in Neural Information Processing Systems},
  volume={36},
  pages={40294--40319},
  year={2023}
}

@inproceedings{li2024garment,
  title={Garment recovery with shape and deformation priors},
  author={Li, Ren and Dumery, Corentin and Guillard, Beno{\^\i}t and Fua, Pascal},
  booktitle={Proceedings of the IEEE/CVF Conference on Computer Vision and Pattern Recognition},
  pages={1586--1595},
  year={2024}
}

@inproceedings{nakayama2025aipparel,
  title={AIpparel: A Multimodal Foundation Model for Digital Garments},
  author={Nakayama, Kiyohiro and Ackermann, Jan and Kesdogan, Timur Levent and Zheng, Yang and Korosteleva, Maria and Sorkine-Hornung, Olga and Guibas, Leonidas J and Yang, Guandao and Wetzstein, Gordon},
  booktitle={Proceedings of the Computer Vision and Pattern Recognition Conference},
  pages={8138--8149},
  year={2025}
}

@article{wang2023prolificdreamer,
  title={Prolificdreamer: High-fidelity and diverse text-to-3d generation with variational score distillation},
  author={Wang, Zhengyi and Lu, Cheng and Wang, Yikai and Bao, Fan and Li, Chongxuan and Su, Hang and Zhu, Jun},
  journal={Advances in neural information processing systems},
  volume={36},
  pages={8406--8441},
  year={2023}
}

@inproceedings{li2025meshpad,
  title={MeshPad: Interactive Sketch-Conditioned Artist-Reminiscent Mesh Generation and Editing},
  author={Li, Haoxuan and Erko{\c{c}}, Ziya and Li, Lei and Sirigatti, Daniele and Rosov, Vladislav and Dai, Angela and Nie{\ss}ner, Matthias},
  booktitle={Proceedings of the IEEE/CVF International Conference on Computer Vision},
  pages={16227--16237},
  year={2025}
}

@article{liu2023modeling,
  title={Modeling realistic clothing from a single image under normal guide},
  author={Liu, Xinqi and Li, Jituo and Lu, Guodong},
  journal={IEEE Transactions on Visualization and Computer Graphics},
  volume={30},
  number={7},
  pages={3995--4007},
  year={2023},
  publisher={IEEE}
}

@article{kwon2023deepiron,
  title={DeepIron: Predicting Unwarped Garment Texture from a Single Image},
  author={Kwon, Hyun-Song and Lee, Sung-Hee},
  journal={arXiv preprint arXiv:2310.15447},
  year={2023}
}

@article{liu2023towards,
  title={Towards garment sewing pattern reconstruction from a single image},
  author={Liu, Lijuan and Xu, Xiangyu and Lin, Zhijie and Liang, Jiabin and Yan, Shuicheng},
  journal={ACM Transactions on Graphics (TOG)},
  volume={42},
  number={6},
  pages={1--15},
  year={2023},
  publisher={ACM New York, NY, USA}
}

@inproceedings{korosteleva2024garmentcodedata,
  title={GarmentCodeData: A dataset of 3D made-to-measure garments with sewing patterns},
  author={Korosteleva, Maria and Kesdogan, Timur Levent and Kemper, Fabian and Wenninger, Stephan and Koller, Jasmin and Zhang, Yuhan and Botsch, Mario and Sorkine-Hornung, Olga},
  booktitle={European Conference on Computer Vision},
  pages={110--127},
  year={2024},
  organization={Springer}
}

@inproceedings{wu2025real,
  title={Real-Time Per-Garment Virtual Try-On with Temporal Consistency for Loose-Fitting Garments},
  author={Wu, Zaiqiang and Shen, I-Chao and Igarashi, Takeo},
  booktitle={Computer Graphics Forum},
  volume={44},
  number={7},
  pages={e70272},
  year={2025},
  organization={Wiley Online Library}
}

@inproceedings{li2024diffavatar,
  title={Diffavatar: Simulation-ready garment optimization with differentiable simulation},
  author={Li, Yifei and Chen, Hsiao-yu and Larionov, Egor and Sarafianos, Nikolaos and Matusik, Wojciech and Stuyck, Tuur},
  booktitle={Proceedings of the IEEE/CVF Conference on Computer Vision and Pattern Recognition},
  pages={4368--4378},
  year={2024}
}

@inproceedings{liu2025multimodal,
  title={Multimodal latent diffusion model for complex sewing pattern generation},
  author={Liu, Shengqi and Cheng, Yuhao and Chen, Zhuo and Ren, Xingyu and Zhu, Wenhan and Li, Lincheng and Bi, Mengxiao and Yang, Xiaokang and Yan, Yichao},
  booktitle={Proceedings of the IEEE/CVF International Conference on Computer Vision},
  pages={17640--17650},
  year={2025}
}

@inproceedings{li2025single,
  title={Single View Garment Reconstruction Using Diffusion Mapping Via Pattern Coordinates},
  author={Li, Ren and Cao, Cong and Dumery, Corentin and You, Yingxuan and Li, Hao and Fua, Pascal},
  booktitle={Proceedings of the Special Interest Group on Computer Graphics and Interactive Techniques Conference Conference Papers},
  pages={1--11},
  year={2025}
}

@article{korosteleva2021generating,
  title={Generating datasets of 3d garments with sewing patterns},
  author={Korosteleva, Maria and Lee, Sung-Hee},
  journal={arXiv preprint arXiv:2109.05633},
  year={2021}
}

@inproceedings{liu2025hybrid,
  title={Hybrid Global-Local Representation with Augmented Spatial Guidance for Zero-Shot Referring Image Segmentation},
  author={Liu, Ting and Li, Siyuan},
  booktitle={Proceedings of the Computer Vision and Pattern Recognition Conference},
  pages={29634--29643},
  year={2025}
}

@misc{lai2025hunyuan3d25highfidelity3d,
      title={Hunyuan3D 2.5: Towards High-Fidelity 3D Assets Generation with Ultimate Details}, 
      author={Tencent Hunyuan3D Team},
      year={2025},
      eprint={2506.16504},
      archivePrefix={arXiv},
      primaryClass={cs.CV},
      url={https://arxiv.org/abs/2506.16504}, 
}

@misc{hunyuan3d22025tencent,
    title={Hunyuan3D 2.0: Scaling Diffusion Models for High Resolution Textured 3D Assets Generation},
    author={Tencent Hunyuan3D Team},
    year={2025},
    eprint={2501.12202},
    archivePrefix={arXiv},
    primaryClass={cs.CV}
}

@article{yang2025sam3dbody,
  title={SAM 3D Body: Robust Full-Body Human Mesh Recovery},
  author={Yang, Xitong and Kukreja, Devansh and Pinkus, Don and Sagar, Anushka and Fan, Taosha and Park, Jinhyung and Shin, Soyong and Cao, Jinkun and Liu, Jiawei and Ugrinovic, Nicolas and Feiszli, Matt and Malik, Jitendra and Dollar, Piotr and Kitani, Kris},
  journal={arXiv preprint; identifier to be added},
  year={2025}
}

@inproceedings{KorostelevaGarmentData,
 author = {Korosteleva, Maria and Lee, Sung-Hee},
 booktitle = {Proceedings of the Neural Information Processing Systems Track on Datasets and Benchmarks},
 editor = {J. Vanschoren and S. Yeung},
 pages = {},
 title = {Generating Datasets of 3D Garments with Sewing Patterns},
 url = {https://datasets-benchmarks-proceedings.neurips.cc/paper/2021/file/013d407166ec4fa56eb1e1f8cbe183b9-Paper-round1.pdf},
 volume = {1},
 year = {2021}
}

@article{team2024gemini,
  title={Gemini: A family of highly capable multimodal models, 2024},
  author={Team, Gemini and Anil, R and Borgeaud, S and Wu, Y and Alayrac, JB and Yu, J and Soricut, R and Schalkwyk, J and Dai, AM and Hauth, A and others},
  journal={arXiv preprint arXiv:2312.11805},
  volume={10},
  year={2024}
}

\end{document}